\documentclass{article} 
\usepackage{iclr2022_conference,times}

\iclrfinalcopy

\usepackage{hyperref}
\usepackage{url}

\usepackage{amsfonts}       
\usepackage{nicefrac}       
\usepackage{microtype}      
\usepackage{xcolor}         
\usepackage{amsmath}
\usepackage{amssymb}
\usepackage{amsthm}

\usepackage{subcaption}
\usepackage{float}
\usepackage{epsfig}
\usepackage{bbm}
\usepackage{comment}
\usepackage{booktabs}
\usepackage{color}
\usepackage{mathrsfs}
\usepackage{wrapfig}
\usepackage{multicol}
\usepackage{multirow}

\usepackage[ruled,vlined,linesnumbered,noresetcount]{algorithm2e}
\usepackage{algpseudocode}

\newtheorem{theorem}{Theorem}

\newtheorem{lemma}[theorem]{Lemma}

\newcommand{\eps}{\varepsilon}

\newcommand{\D}{\mathcal{D}}
\newcommand{\E}{\mathbb{E}}

\newcommand{\R}{\mathbb{R}}

\usepackage{bbm}

\DeclareMathOperator*{\argmin}{\arg\!\min}
\DeclareMathOperator*{\argmax}{\arg\!\max}

\newcommand{\trainbudget}{m_{\text{train}}}
\newcommand{\evalbudget}{m_{\text{pred}}}
\newcommand{\utilityLearningAlg}{\mathcal{A}}

\newcommand{\heuristicSampler}{\mathcal{V}_{\text{samp}}}
\newcommand{\heuristicEstimator}{\mathcal{V}_{\text{est}}}
\newcommand{\trainSample}{\mathcal{S}_{\text{train}}}
\newcommand{\predSample}{\mathcal{S}_{\text{pred}}}

\newcommand{\uf}{v}
\newcommand{\duf}{\hat v}

\newcommand{\estar}{e^{\star}}
\newcommand{\equi}{\Leftrightarrow}
\newcommand{\hybridduf}{\Tilde{v}}
\newcommand{\norm}[1]{\left\lVert#1\right\rVert}
\newcommand{\mperm}{m_{\text{perm}}}

\title{Improving Cooperative Game Theory-based Data Valuation via Data Utility Learning}

\author{Tianhao Wang \\
Princeton University\\
\texttt{tianhaowang@princeton.edu} \\
\And
Yu Yang\\
Xi'an Jiaotong University \\
\texttt{yangyu21@xjtu.edu.cn} \\
\And
Ruoxi Jia \\
Virginia Tech \\
\texttt{ruoxijia@vt.edu}
}

\begin{document}

\setlength{\abovedisplayskip}{5pt}
\setlength{\belowdisplayskip}{5pt}

\maketitle

\begin{abstract}
The Shapley value (SV) and Least core (LC) are classic methods in cooperative game theory for cost/profit sharing problems. Both methods have recently been proposed as a principled solution for \emph{data valuation} tasks, i.e., quantifying the contribution of individual datum in machine learning. However, both SV and LC suffer computational challenges due to the need for retraining models on combinatorially many data subsets. In this work, we propose to boost the efficiency in computing Shapley value or Least core by learning to estimate the performance of a learning algorithm on unseen data combinations. Theoretically, we derive bounds relating the error in the predicted learning performance to the approximation error in SV and LC. Empirically, we show that the proposed method can significantly improve the accuracy of SV and LC estimation.

\end{abstract}

\vspace{-7mm}
\section{Introduction}
\label{sec:intro}

\vspace{-4mm}
Data valuation aims to quantify the usefulness of each data source for machine learning (ML) tasks, which has many use cases and has gained lots of attention recently. For instance, it can inform the implementation of policies that enable individuals to control how their data is used and monetized by third parties \citep{voigt2017eu}. Moreover, data value allows users to filter out poor quality data and identify data that are important to collect in the future \citep{jia2019scalability,ghorbani2019data}. Existing data valuation approaches leverage solution concepts from cooperative game theory, specifically the Shapley value (SV) and Least core (LC), as a fair notion for data value. However, computing or even approximating these data value notions requires evaluating the learning performance (i.e., retraining) on many different combinations of data sources, which could be very computationally expensive. In this work, we develop a general framework to improve the effectiveness of sampling-based SV or LC estimation heuristics. Specifically, we propose to learn to predict the performance of a learning algorithm for an input dataset (which we refer to as \emph{data utility learning}) and use the trained predictor to estimate the learning performance on a dataset without retraining. As the predicted learning performance may be noisy, we derive the estimation guarantee of SV/LC under a hybrid of clean and noisy samples. Moreover, we conduct extensive experiments and show that data utility learning can significantly improve SV and LC approximation accuracy. 

\vspace{-3mm}
\section{Background and Related Work}
\label{sec:related-work}

\vspace{-4mm}
The goal of data valuation is to quantify the contribution of each training data point to a learning task.
Game-theoretic formulations of data valuation have become popular recently. 
Formally, a cooperative game considers a set of \emph{players} $N = \{1, \ldots, n\}$, and a \emph{characteristic function} $v: 2^N \rightarrow \R$ assigns a value to every subset (often referred to as a \emph{coalition}) $S \subseteq N$. 
In the context of data valuation, $N$ corresponds to the dataset and each player $i \in N$ is a data point; the characteristic function $v$ takes a subset of data points as input, and output the performance score (e.g., test accuracy) of a learning algorithm trained on the data subset. 
We will refer to $v$ as \emph{data utility function} in data valuation context, which stresses that $v$ reflects the utility of the input dataset for the ML algorithm. 
Shapley value has been widely used as a data value notion~\citep{ghorbani2019data, jia2019towards, jia2019efficient, jia2019scalability, wang2020principled}, as it uniquely satisfies a set of desirable properties for profit sharing problems. The Shapley value $\phi(v)_i$ for data point $i$ is defined as
\begin{align}
   \phi(v)_i = \frac{1}{n} \sum_{S\subseteq N\setminus\{i\}} \frac{1}{{n-1 \choose |S|}}
\left[v(S\cup \{i\})-v(S)\right]
\label{eq:shapley}
\end{align}
In a word, Shapley value is a weighted average of the marginal contributions of a data point over all possible subsets. 
Recently, \cite{yan2020ifyoulike} propose to use the Least core, another classic solution concept in cooperative game theory, as an alternative to Shapley value for data valuation. 
Least core vector $\psi \in \R^n$ minimizes the excess of the actual utility over the assigned value for every possible subset and is computed by solving the linear programming problem below:
\begin{align}
\min_{\psi} e \quad \text{s.t. }\sum_{i=1}^n \psi_i = v(N), \sum_{i \in S}\psi_i +e \geq v(S), \forall S \subseteq N 
\label{eq:leastcore}
\end{align}


\vspace{-1mm}
The fairness properties of SV and LC provide strong motivation for using them in data valuation. We point readers to the cited references for a detailed discussion on it. However, from the definitions in (\ref{eq:shapley}) and (\ref{eq:leastcore}), it can be seen that the exact calculation of both SV and LC requires evaluating the data utility function, i.e., retraining, on every possible subset of the training data, which is $2^n$ in total. Indeed, the exact computation of SV and LC is NP-hard in general \citep{deng1994complexity, faigle2000note}, which limits their applicability in real-world data valuation applications even at the scale of hundreds of data points. Several approximation heuristics, such as TMC-Shapley \citep{ghorbani2019data} and KNN-Shapley \citep{jia2019efficient}, have been proposed to approximate Shapley value. Despite their computational advantage, they are biased in nature, limiting their applicability to sensitive applications such as designing monetary rewards for data sharing or assigning responsibility for ML decisions. On the other hand, unbiased Shapley estimators such as Permutation Sampling \citep{castro2009polynomial} and Group Testing \citep{jia2019towards} still require a large number of learning performance evaluations for any descent approximation accuracy. 

\vspace{-1mm}
The idea of learning characteristic function was previously proposed in \cite{balcan2015learning}, and they studied the PAC learnability of several well-known classes of cooperative games. \cite{yan2020evaluating} is the closest work to our paper, where they use linear regression with interaction terms to learn characteristic functions and use the learned linear model to estimate SV. However, their technique cannot be directly applicable to data valuation due to many reasons, which we discuss in Appendix \ref{appendix:CGA}.

\vspace{-5mm}
\section{Data Valuation via Data Utility Learning}
\label{sec:dul-to-datavaluation}

\vspace{-4mm}
The major idea underlying the existing data valuation heuristics is to evaluate the data utility only on some sampled subsets and then estimate the data value based on the samples of subset-utility pairs (i.e., Monte Carlo or its variants). We will refer to the samples of subset-utility pairs as \emph{utility samples}. 
The key idea of our approach is that, with the utility samples, we can potentially use a parametric model $\duf$ to learn and approximate the data utility function $v$. We can then use $\duf$ to predict the utility for additional subsets that are not sampled previously. 

\vspace{-3mm}
\paragraph{Sampling-based SV/LC Estimation Heuristics. }
We first build an abstraction for sampling-based data value approximation heuristics. Sampling-based heuristics comprise of all the existing \emph{unbiased} heuristics~\citep{castro2009polynomial, jia2019towards, yan2020ifyoulike} as well as some of the biased heuristics (e.g., TMC Shapley \citep{ghorbani2019data}). 
A sampling-based heuristics can be characterized by two components: a sampler $\heuristicSampler$ and an estimator $\heuristicEstimator$. 
Under some \emph{training budget} $\trainbudget > 0$, the heuristic sampler $\heuristicSampler$ takes a dataset $N$ and outputs a set of utility samples $\{(S_i, v(S_i))\}_{i=1}^{\trainbudget}$ where each $S_i \subseteq N$ is sampled according to certain distributions. $v(S_i)$ is usually obtained by retraining models on $S_i$ and obtain the performance scores of trained model. 
The heuristic estimator $\heuristicEstimator$ then takes the utility samples and computes the estimation of the corresponding solution concept (i.e., Shapley or Least core). 
\textbf{Example-1: Permutation Sampling estimator for Shapley value.} The definition of Shapley value in (\ref{eq:shapley}) can also be conveniently expressed in terms of permutations
$
\phi(v)_i = \frac{1}{n!} \sum_{\pi \in \Pi(N)} \left[ v(P_i^{\pi} \cup \{i\}) - v(P_i^{\pi}) \right]
$
where $\Pi(N)$ is the collection of $n!$ permutations of $N$, and $P_i^{\pi}$ represents the set of data points ranked lower than $i$ in the permutation $\pi$. 
Since this is just the expectation of $v(P_i^{\pi} \cup \{i\}) - v(P_i^{\pi})$ when $\pi$ is uniformly sampled from $\Pi(N)$, an obvious approximation is the simple Monte Carlo estimator
\begin{align}
    \hat \phi(v)_i = \frac{1}{\mperm} \sum_{j=1}^{\mperm} \left[ v(P_i^{\pi_j} \cup \{i\}) - v(P_i^{\pi_j}) \right] \label{eq:perm-mc}
\end{align}
for a uniform of samples $\pi_1, \ldots, \pi_{\mperm} \in \Pi(N)$. 
In this case, $\heuristicSampler$ samples $\pi_1, \ldots, \pi_{\mperm} \in \Pi(N)$ and compute $v(P_i^{\pi_j} \cup \{i\})$ for each $i \in N$ (so the sampling budget $\trainbudget \ge n \mperm$ as every permutation requires $n$ times model retraining). $\heuristicEstimator$ takes utility samples and estimate SV according to (\ref{eq:perm-mc}) for each $i \in N$. 
\textbf{Example-2: Monte Carlo estimator for Least Core.}
Given $\trainbudget$ utility samples $\{(N, \uf(N))\} \bigcup \{(S_j, v(S_j))\}_{j=1}^{\trainbudget-1}$ where each $S_j \subset N$, 
it is straightforward to give a Monte Carlo estimator for Least core in (\ref{eq:leastcore}) by solving the following linear program:
\begin{align}
\min_{\psi} e \quad \text{s.t. }\sum_{i=1}^n \psi_i = v(N), \sum_{i \in S_j}\psi_i +e \geq v(S_j), j = 1, \ldots, \trainbudget-1
\label{eq:leastcore-mc}
\end{align}
In this case, $\heuristicSampler$ compute $v(N)$ as well as $\{(S_j, v(S_j))\}_{j=1}^{\trainbudget-1}$ where $S_j \subset N$ are sampled uniformly at random, and $\heuristicEstimator$ solves the linear program in (\ref{eq:leastcore-mc}). 

\vspace{-3mm}
\paragraph{Boosting Sampling-based Heuristics via Utility Function Learning. }
Algorithm \ref{alg:data-value-with-dul} summarizes our algorithm for accelerating data valuation with data utility learning. Our algorithm leverages the set of utility samples $\trainSample = \{(S_i, v(S_i))\}_{i=1}^{\trainbudget}$ that need to be sampled in the existing heuristics, and uses it as the training data to learn data utility function $\uf$. 
In our experiment, we use a simple neural network to train on $\trainSample$, where each $S_i$ is encoded as a $n$-dimensional binary vector whose entry indicates the existence of a corresponding data point. 
Once we obtain the learned data utility model $\duf$, we can use it to predict the utilities for much more subsets and obtain additional samples $\predSample = \{(S_i, \duf(S_i))\}_{i=1}^{\evalbudget}$, where we call $\evalbudget$ the \emph{prediction budget}. We sample the additional subsets with the \emph{same} distribution followed by the heuristic sampler $\heuristicSampler$. 
At last, we feed the combination of the original and predicted utility samples $\trainSample \cup \predSample$ to $\heuristicEstimator$. Generally, querying $\duf$ is usually far more efficient than retraining the ML model. Therefore, the additional predicted utility samples $\predSample$ can almost be acquired for free! In practice, we could set $\evalbudget \gg \trainbudget$. 

\begin{algorithm}[tb]
\setlength{\textfloatsep}{-10pt}
\SetAlgoLined
\SetKwInOut{Input}{input}
\SetKwInOut{Output}{output}
\Input{Data valuation heuristic $(\heuristicSampler, \heuristicEstimator)$, Dataset $N = \{1, \ldots, n\}$, 
Training budget $\trainbudget$, Prediction budget $\evalbudget$, Learning Algorithm for Data Utility Model $\utilityLearningAlg$.}
\Output{Estimated solution concept $x \in \mathbb{R}^n$.}

Sample $\trainbudget$ data subsets and their utilities $\trainSample = \{(S_i, v(S_i))\}_{i=1}^{\trainbudget}$ with sampler $\heuristicSampler$.

Train data utility model $\duf \leftarrow \utilityLearningAlg(\trainSample)$.

Sample $\evalbudget$ additional data subsets $\{S_i\}$ according to $\heuristicSampler$, and obtain their \emph{predicted} utilities $\predSample = \{(S_i, \duf(S_i))\}_{i=1}^{\evalbudget}$. 

Compute the corresponding solution concept $x \leftarrow \heuristicEstimator \left(\trainSample \cup \predSample \right)$. 

\Return{$x$}
\caption{Algorithm for Accelerating The Estimation of Data Valuation Solution Concepts}
\label{alg:data-value-with-dul}
\end{algorithm}

\vspace{-5mm}
\section{Theoretical Analysis}
\label{sec:theory}

\vspace{-3mm}
Since the trained parametric function $\duf$ may not fully recover $\uf$, we investigate the reliability of SV and LC estimated from a hybrid of $\trainbudget$ clean samples from $\uf$ and $\evalbudget$ noisy samples from $\duf$. 
We denote the ratio of noisy samples as $\gamma = \evalbudget / m$. 
Due to space constraint, we defer the worst-case, information-theoretic result to Appendix \ref{appendix:sv-infotheory} where the noise $\duf - \uf$ can be adversarially distributed. 
Below, we show the average-case guarantee in SV estimation using Permutation Sampling \citep{castro2009polynomial} with hybrid utility samples. 
Different from the worst-case analysis in Appendix \ref{appendix:sv-infotheory}, we consider the scenario when error $\duf - \uf$ is drawn from a \emph{smooth} distribution, a commonly used assumption in sensitivity analysis \citep{yan2020evaluating, gupta2017pac} and is arguably reasonable when $\duf$ is a trained on uniformly distributed samples. 
We say $\hat \phi$ is an $(\eps, \delta)$-approximation to the true SV in $p$-norm if $\Pr_{\hat \phi} \left[ \norm{\phi - \hat \phi}_p \le \eps \right] \ge 1-\delta$. 
By Hoeffding, when $p=1$ and the samples are all clean ($\gamma=0$), we can achieve $(\eps, \delta)$-approximation with $m = O\left( \frac{n^2}{\eps^2} \log (\frac{n}{\delta}) \right)$ samples. 

\begin{theorem}
\label{thm:sv-perm}
If there exists two functions $\kappa_0, \kappa_1$ s.t. the random noise $\duf - \uf \sim \D_n$ has conditional distribution s.t. $\kappa_{0}(r) \leq$ $\operatorname{Pr}_{\mathcal{D}_{n}}\left(x \mid\|x\|_{1}=r\right) \leq \kappa_{1}(r)$ for all $r \ge 0$ and $x$ in $\mathcal{D}_{n}$'s support, then with sample $m = O\left( \frac{n^2}{\eps^2} \log (\frac{n}{\delta}) \right)$ and noise ratio $\gamma$, permutation sampling achieves $(\eps + \gamma cn, \delta)$-approximate Shapley value in $\ell_1$ norm, where $c = \frac{1}{2^{n-1}} \E_r\left[ \frac{\kappa_1(r)}{\kappa_0(r)}r \right]$. 
\end{theorem}
The above result says that when the conditional error distribution $\duf - \uf$ is not very ``peaky'' anywhere over all noise $\duf - \uf$ of the same magnitude, 
then the sample complexity of permutation sampling with hybrid utility samples is the same as regular permutation sampling, except that it has an extra irreducible error term $\gamma cn$. As long as the error distribution is fairly smooth, i.e., $\kappa_1(r) / \kappa_0(r)$ is close to 1 for all $r$, this error term will be neglectable due to the $2^n$ in the denominator.

We also derive the guarantee for Least core estimation with Monte Carlo approach in (\ref{eq:leastcore-mc}). We follow \cite{yan2020ifyoulike} and define the $(\eps, \delta)$-probably approximate least core to be the vector $\psi \in \R^n$ s.t. $\Pr_{S \sim \D} \left[ \sum_{i \in S} \psi_{i}+\estar+\eps \geq \uf(S) \right] \geq 1-\delta$ where $\estar$ is the minimized objective solved from the exact formula (\ref{eq:leastcore}), so it is unknown in practice. Nevertheless, we can still derive meaningful guarantee for the estimated $\psi$ in terms of this unknown $\estar$. 

\begin{theorem}
\label{thm:least-core}
Given distribution $\D$ over $2^N$, and $\delta, \Delta, \eps > 0$, if for every $S \subseteq N$, the error in $\hat v(S)$ has property $(1-\eps \sigma) v(S) \le \hat v(S) \le (1+\eps \sigma) v(S)$ with probability at least $1-\mu$ over the learning process of $\hat v$, then solving the linear program in (\ref{eq:leastcore-mc}) over 
$
O\left(\frac{\tau^2\left(\log n+\log \left(1 / \Delta\right)\right)}{\eps^{2} \delta^{2}}\right)
$
coalitions sampled from $\D$ with noisy sample ratio $\gamma$, where $\tau = \frac{\max_{S} v(S) + \eps \sigma}{\min_{S \neq \emptyset} v(S)}$, gives $(\eps (1+\gamma \sigma), \delta + \gamma \sigma)$-probably approximately least core with probability at least $1-\Delta - \evalbudget \mu$.
\end{theorem}
When $\sigma = 0$, Theorem \ref{thm:least-core} recovers Theorem 2 in \cite{yan2020ifyoulike} for the case of $\duf = \uf$. This result suggests that with number of samples polynomial in $O(\log n)$, one can still obtain a good approximation of least core with some additional irreducible error due to the error in $\duf$.

\vspace{-5mm}
\section{Evaluation}
\label{sec:eval}

\vspace{-3mm}

\paragraph{Settings.}
We first assess the performance of Shapley and Least core estimators on small enough datasets (\# data $\le 15$); in this case, we are able to directly calculate the true data value with the exact formula in (\ref{eq:shapley}) and (\ref{eq:leastcore}), then evaluate and compare the estimation error of different heuristics. 
Due to space constraint, we defer the results for larger datasets and data groups to the Appendix. 
\textbf{Baselines:} For Shapley value estimation, we consider the two existing unbiased estimators as our baselines: \textbf{(1) Permutation Sampling (Perm)} \citep{castro2009polynomial}, \textbf{(2) Group Testing (GT)} \citep{jia2019towards}, which is an improved Monte Carlo algorithm based on group testing theory. We also compare with \textbf{CGA} approach \citep{yan2020evaluating} mentioned in related work, which approximate $\uf$ with a linear regression model with interaction terms. 
For LC estimation, we use \textbf{Monte Carlo (MC)} approach \citep{yan2020ifyoulike} as the baseline. For all experiments, $\duf$ is a 3-layer MLP model, and the input data subsets are encoded as binary vector.

\newcommand{\width}{0.6}
\begin{wrapfigure}{R}{\width \textwidth}
    \vspace{-5.9mm}
    \setlength\belowcaptionskip{-10pt}
    \centering
    \includegraphics[width=\width \textwidth]{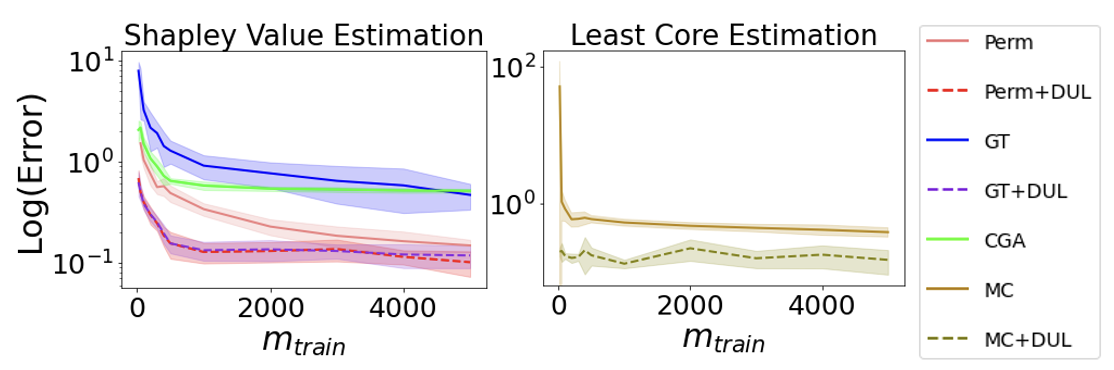}
    \caption{$\ell_1$ approximation error for Shapley and Least core values with the change of number of sampled utility scores. 
    `DUL' stands for data utility learning.
    }
    \label{fig:error-simulation}
\end{wrapfigure}

\vspace{-3mm}
\paragraph{Results. }

We test the performance of different SV/LC estimation heuristics on tiny datasets ($<15$ data points) so that it is computationally feasible to compute the \emph{exact} SV and LC, and we can directly calculate the estimation error. 
We randomly sample 15 data points from the famous Iris dataset \citep{pedregosa2011scikit}. A support-vector machine (SVM) classifier trained on the 15 data points achieves around 94\% test accuracy. 
After training the utility model, we obtain the estimated data utilities for all the subsets not sampled before ($\evalbudget = 2^{15} - \trainbudget$). We then estimate SV and LC using the exact calculation formula, with the hybrid of clean and noisy data utilities. 
We show $\ell_1$ estimation errors in Figure \ref{fig:error-simulation}, and defer the results for $\ell_2$ and $\ell_\infty$ errors to Appendix\footnote{It is easy to see that LC may not be unique. In the experiment, when we talk about the LC, we always refer to the vector $\psi$ that has the smallest $\ell_2$ norm, following the tie-breaking rule in \cite{yan2020ifyoulike}. It is difficult to directly use $(\eps, \delta)$-LC definition in Section \ref{sec:theory} to compare LC estimation as it has two error dimensions. }. As we can see, with a relatively small $\trainbudget$ (e.g., 500), data utility learning can significantly reduce the estimation errors for both SV and LC. Utility prediction per se may introduce additional computational costs; yet, these costs are often negligible compared to model retraining. Moreover, CGA-based SV performs poorly, because there are high-order data interactions in data utility functions which CGA cannot successfully capture, as discussed in Appendix \ref{appendix:CGA}.

\vspace{-5mm}
\section{Conclusion}
\label{sec:conclusion}

\vspace{-3mm}
This work proposes a generic framework that can significantly improve the accuracy of all the existing unbiased SV/LC estimation methods through learning data utility functions. Particularly, our approach can be extended to other cooperative games where querying characteristic functions are expensive, e.g., bidders' valuation functions in combinatorial auctions \citep{lehmann2006combinatorial}. With this paper, we march one step closer towards making data valuation practical and hope to inspire more research in this direction amongst the ML community.

\newpage

\bibliography{ref}

\begin{thebibliography}{20}
\providecommand{\natexlab}[1]{#1}
\providecommand{\url}[1]{\texttt{#1}}
\expandafter\ifx\csname urlstyle\endcsname\relax
  \providecommand{\doi}[1]{doi: #1}\else
  \providecommand{\doi}{doi: \begingroup \urlstyle{rm}\Url}\fi

\bibitem[Balcan et~al.(2015)Balcan, Procaccia, and Zick]{balcan2015learning}
Maria~Florina Balcan, Ariel~D Procaccia, and Yair Zick.
\newblock Learning cooperative games.
\newblock In \emph{Twenty-Fourth International Joint Conference on Artificial
  Intelligence}, 2015.

\bibitem[Balkanski et~al.(2017)Balkanski, Syed, and
  Vassilvitskii]{balkanski2017statistical}
Eric Balkanski, Umar Syed, and Sergei Vassilvitskii.
\newblock Statistical cost sharing.
\newblock \emph{Advances in Neural Information Processing Systems}, 30, 2017.

\bibitem[Castro et~al.(2009)Castro, G{\'o}mez, and
  Tejada]{castro2009polynomial}
Javier Castro, Daniel G{\'o}mez, and Juan Tejada.
\newblock Polynomial calculation of the shapley value based on sampling.
\newblock \emph{Computers \& Operations Research}, 36\penalty0 (5):\penalty0
  1726--1730, 2009.

\bibitem[Deng \& Papadimitriou(1994)Deng and Papadimitriou]{deng1994complexity}
Xiaotie Deng and Christos~H Papadimitriou.
\newblock On the complexity of cooperative solution concepts.
\newblock \emph{Mathematics of operations research}, 19\penalty0 (2):\penalty0
  257--266, 1994.

\bibitem[Dua \& Graff(2017)Dua and Graff]{adultdata}
Dheeru Dua and Casey Graff.
\newblock {UCI} machine learning repository, 2017.
\newblock URL \url{http://archive.ics.uci.edu/ml}.

\bibitem[Faigle et~al.(2000)Faigle, Kern, and Paulusma]{faigle2000note}
Ulrich Faigle, Walter Kern, and Dani{\"e}l Paulusma.
\newblock Note on the computational complexity of least core concepts for
  min-cost spanning tree games.
\newblock \emph{Mathematical methods of operations research}, 52\penalty0
  (1):\penalty0 23--38, 2000.

\bibitem[Ghorbani \& Zou(2019)Ghorbani and Zou]{ghorbani2019data}
Amirata Ghorbani and James Zou.
\newblock Data shapley: Equitable valuation of data for machine learning.
\newblock In \emph{International Conference on Machine Learning}, pp.\
  2242--2251. PMLR, 2019.

\bibitem[Gupta \& Roughgarden(2017)Gupta and Roughgarden]{gupta2017pac}
Rishi Gupta and Tim Roughgarden.
\newblock A pac approach to application-specific algorithm selection.
\newblock \emph{SIAM Journal on Computing}, 46\penalty0 (3):\penalty0
  992--1017, 2017.

\bibitem[Jia et~al.(2019{\natexlab{a}})Jia, Dao, Wang, Hubis, Gurel, Li, Zhang,
  Spanos, and Song]{jia2019efficient}
Ruoxi Jia, David Dao, Boxin Wang, Frances~Ann Hubis, Nezihe~Merve Gurel, Bo~Li,
  Ce~Zhang, Costas~J Spanos, and Dawn Song.
\newblock Efficient task-specific data valuation for nearest neighbor
  algorithms.
\newblock \emph{arXiv preprint arXiv:1908.08619}, 2019{\natexlab{a}}.

\bibitem[Jia et~al.(2019{\natexlab{b}})Jia, Dao, Wang, Hubis, Hynes, G{\"u}rel,
  Li, Zhang, Song, and Spanos]{jia2019towards}
Ruoxi Jia, David Dao, Boxin Wang, Frances~Ann Hubis, Nick Hynes, Nezihe~Merve
  G{\"u}rel, Bo~Li, Ce~Zhang, Dawn Song, and Costas~J Spanos.
\newblock Towards efficient data valuation based on the shapley value.
\newblock In \emph{The 22nd International Conference on Artificial Intelligence
  and Statistics}, pp.\  1167--1176. PMLR, 2019{\natexlab{b}}.

\bibitem[Jia et~al.(2019{\natexlab{c}})Jia, Wu, Sun, Xu, Dao, Kailkhura, Zhang,
  Li, and Song]{jia2019scalability}
Ruoxi Jia, Fan Wu, Xuehui Sun, Jiacen Xu, David Dao, Bhavya Kailkhura,
  Ce~Zhang, Bo~Li, and Dawn Song.
\newblock Scalability vs. utility: Do we have to sacrifice one for the other in
  data importance quantification?
\newblock \emph{arXiv preprint arXiv:1911.07128}, 2019{\natexlab{c}}.

\bibitem[Lehmann et~al.(2006)Lehmann, Lehmann, and
  Nisan]{lehmann2006combinatorial}
Benny Lehmann, Daniel Lehmann, and Noam Nisan.
\newblock Combinatorial auctions with decreasing marginal utilities.
\newblock \emph{Games and Economic Behavior}, 55\penalty0 (2):\penalty0
  270--296, 2006.

\bibitem[Pedregosa et~al.(2011)Pedregosa, Varoquaux, Gramfort, Michel, Thirion,
  Grisel, Blondel, Prettenhofer, Weiss, Dubourg, et~al.]{pedregosa2011scikit}
Fabian Pedregosa, Ga{\"e}l Varoquaux, Alexandre Gramfort, Vincent Michel,
  Bertrand Thirion, Olivier Grisel, Mathieu Blondel, Peter Prettenhofer, Ron
  Weiss, Vincent Dubourg, et~al.
\newblock Scikit-learn: Machine learning in python.
\newblock \emph{the Journal of machine Learning research}, 12:\penalty0
  2825--2830, 2011.

\bibitem[Pinto et~al.(2011)Pinto, Stone, Zickler, and Cox]{pinto2011scaling}
Nicolas Pinto, Zak Stone, Todd Zickler, and David Cox.
\newblock Scaling up biologically-inspired computer vision: A case study in
  unconstrained face recognition on facebook.
\newblock In \emph{CVPR 2011 WORKSHOPS}, pp.\  35--42. IEEE, 2011.

\bibitem[Shalev-Shwartz \& Ben-David(2014)Shalev-Shwartz and
  Ben-David]{shalev2014understanding}
Shai Shalev-Shwartz and Shai Ben-David.
\newblock \emph{Understanding machine learning: From theory to algorithms}.
\newblock Cambridge university press, 2014.

\bibitem[Srivastava et~al.(2014)Srivastava, Hinton, Krizhevsky, Sutskever, and
  Salakhutdinov]{srivastava2014dropout}
Nitish Srivastava, Geoffrey Hinton, Alex Krizhevsky, Ilya Sutskever, and Ruslan
  Salakhutdinov.
\newblock Dropout: a simple way to prevent neural networks from overfitting.
\newblock \emph{The journal of machine learning research}, 15\penalty0
  (1):\penalty0 1929--1958, 2014.

\bibitem[Voigt \& Von~dem Bussche(2017)Voigt and Von~dem Bussche]{voigt2017eu}
Paul Voigt and Axel Von~dem Bussche.
\newblock The eu general data protection regulation (gdpr).
\newblock \emph{A Practical Guide, 1st Ed., Cham: Springer International
  Publishing}, 10:\penalty0 3152676, 2017.

\bibitem[Wang et~al.(2020)Wang, Rausch, Zhang, Jia, and
  Song]{wang2020principled}
Tianhao Wang, Johannes Rausch, Ce~Zhang, Ruoxi Jia, and Dawn Song.
\newblock A principled approach to data valuation for federated learning.
\newblock In \emph{Federated Learning}, pp.\  153--167. Springer, 2020.

\bibitem[Yan \& Procaccia(2020)Yan and Procaccia]{yan2020ifyoulike}
Tom Yan and Ariel~D Procaccia.
\newblock If you like shapley then you’ll love the core, 2020.

\bibitem[Yan et~al.(2020)Yan, Kroer, and Peysakhovich]{yan2020evaluating}
Tom Yan, Christian Kroer, and Alexander Peysakhovich.
\newblock Evaluating and rewarding teamwork using cooperative game
  abstractions.
\newblock \emph{arXiv preprint arXiv:2006.09538}, 2020.

\end{thebibliography}
\bibliographystyle{iclr2022_conference}

\newpage

\appendix

\section{Comparison with Cooperative Game Abstraction (CGA). }
\label{appendix:CGA}

\cite{yan2020evaluating} propose a similar idea which approximates Shapley value through learning the characteristic functions in a cooperative game, where the parametric model they use to learn is called cooperative game abstraction (CGA). CGA is essentially a linear regression model where the variables are small subsets of players. The order of CGA refers to the largest size of player groups included as variables in the linear regression. One advantage of CGA is that it can recover Shapley value directly from its parameters if the function of CGA perfectly matches the characteristic function. Hence, \cite{yan2020evaluating} propose to learn the characteristic function using CGA with certain amount of samples, and then compute Shapley value through CGA's trained parameters. For data valuation problem, the characteristic function is the data utility function $\uf$. In this sense, CGA can be viewed as a special case of data utility learning where the data utility model is linear regression. However, we argue that CGA may not be a suitable model for data utility learning for two main reasons:
(i) CGA is only suitable for certain types of games where the interactions only exist among small groups of players, e.g., the team performance in basketball games. On the contrary, interactions between large groups of data points might still be strong. 
Thus, CGA is not a suitable model for data utility learning in nature, and we confirmed this point in Section \ref{sec:eval}. 
(ii) CGA has a poor scalability even for one with low order, e.g. for a third-order CGA with 2000 players, the number of total parameters is ${2000 \choose 2} + 2000 = 2,001,000$, while the number of total parameters of a $2000\times 512 \times 1$ fully-connected neural network is only around half of it ($1,024,512$). 
We also note that the parameters of CGA only has a closed form for computing Shapley value, but not the Least core.




\section{Information-theoretic Result for SV Estimation}
\label{appendix:sv-infotheory}

For SV, we related the error in $\duf$ to the error in the final estimation. We start with an information-theoretic result when the Shapley value is computed by the exact formula in (\ref{eq:shapley}) with $\trainbudget$ clean samples and $\evalbudget = 2^n - \trainbudget$ noisy samples. 
As $\uf$ and $\duf$ can be viewed as a $2^n$-dimensional vector, the input for (\ref{eq:shapley}) is the hybrid data utility vector $\hybridduf \in \R^{2^n}$ where the entry correspond to subset $S$ is $\uf(S)$ if it is within the clean samples, and $\duf(S)$ otherwise. 
The following result shows that, by smartly allocate the clean sample budget, one can significantly reduce the error in the final estimation.
\begin{theorem}
\label{thm:sv-infotheory}
For any integer $-1 \le p \le \log(n)$, if for all $|S| \le p$ and $|S| \ge n-p$ we have clean samples $\uf(S)$ then the estimation error of SV is bounded by 
$
\norm{ \phi(v) - \phi(\hybridduf) }_2 \le \Tilde{O}\left(\frac{1}{n^{1+p/2}}\right) \norm{ v - \hybridduf }_2
$. 
\end{theorem}
The above results imply that, whenever $\trainbudget \ge 2\sum_{i=0}^p {n \choose i}$ for some $p \ge 0$, one can largely improve the Shapley value estimation guarantee by computing $v(S)$ for $S$ very small or very large size. 
The case of $p = -1$ (i.e., $\hybridduf = \duf$) recovers Theorem 3 in \cite{yan2020evaluating}. In particular, even when $p=0$, i.e., only two clean samples $\uf(\emptyset)$ and $\uf(N)$ are computed, the $\ell_2$ error guarantee can be improved by $O\left(\frac{1}{\sqrt{n}}\right)$. In practice, to learn a good $\duf$ s.t. $\norm{ \uf - \duf }_2$ is small, the clean samples need to be distributed uniformly over $2^N$. However, Theorem \ref{thm:sv-infotheory} still suggests that when there are spare training budget, one can distribute it to very small or very large subsets.

\subsection{Proof}
The proof extends the proof of Theorem 3 in \cite{yan2020evaluating}. Like them, we use the fact that Shapley value is a linear mapping $\phi: \R^{2^n} \rightarrow \R^n$ taking $v$ to $\phi(v)$. 
We can describe this map with a matrix $S_n \in \R^{n \times 2^n}$. 
For any hybrid estimation $\hybridduf$, we have 
\begin{align}
    \norm{\phi(v) - \phi(\hybridduf)} = 
    \norm{S_n v - S_n \hybridduf} = \norm{S_n (v - \hybridduf)}
\end{align}
Suppose there are $\trainbudget$ clean estimations of $v(S)$ and the rest of $2^n - \trainbudget$ are noisy estimation. Then $v - \hybridduf$ has exactly $\trainbudget$ entries are $0$, which means that the corresponding $\trainbudget$ columns of $S_n$ are not used; we use $\Tilde{S}_n \in \R^{n \times (2^n-\trainbudget)}$ to denote the submatrix and $\hybridduf' \in \R^{n \times (2^n-\trainbudget)}$ to denote the compressed vector. 
Thus we have 
\begin{align}
     \norm{S_n (v - \hybridduf)} = \norm{\Tilde{S}_n \hybridduf'}
     \le \norm{\Tilde{S}_n}_{op} \norm{\hybridduf'}
\end{align}
So the question reduce to selecting $2^n-\trainbudget$ columns of $S_n$ and minimize 
\begin{align}
\norm{\Tilde{S}_n}_{op} = \sqrt{\sigma_{\max} ( \Tilde{S}_n^T \Tilde{S}_n ) } = \sqrt{ \sigma_{\max} ( \Tilde{S}_n \Tilde{S}_n^T ) }
\end{align}

We analyze the best column subset selection approach to minimize $\sigma_{\max} \left( \Tilde{S}_n \Tilde{S}_n^T \right)$. Denote the selector vector $\alpha \in \{0, 1\}^{2^n}$, where $\alpha_S = 1$ if the column of $S$ is kept (noisy), and $0$ otherwise (clean). So $\norm{\alpha}_1 = 2^n - \trainbudget$. We denote $\alpha^{(k)} = \left| \{ \alpha_S | |S|=k, \alpha_S=1 \} \right|$, so $\sum_{k=0}^n \alpha^{(k)} = 2^n - \trainbudget$. 

For the $i$th row of $S_n$, $(S_n)_i$, we know that the entry in this row corresponding to subset $S$ is $\frac{1}{n} {n-1 \choose |S|-1}^{-1}$ if $i \in S$, and $-\frac{1}{n} {n-1 \choose |S|}^{-1}$ if $i \notin S$, thus we can compute every diagonal and non-diagonal entry. The $i$th diagonal entry of $\Tilde{S}_n \Tilde{S}_n^T$ is
\begin{align}
    d_{ii}
    &= (\Tilde{S}_n)_i^T (\Tilde{S}_n)_i \\
    &= \sum_{S\in 2^n, i\in S} \alpha_S \left( \frac{1}{n} {n-1 \choose |S|-1}^{-1} \right)^2 
    + \sum_{S\in 2^n, i\notin S} \alpha_S \left( -\frac{1}{n} {n-1 \choose |S|}^{-1} \right)^2
\end{align}
and the $(i, j)$ non-diagonal entry is 
\begin{align}
    d_{ij}
    &= (\Tilde{S}_n)_i^T (\Tilde{S}_n)_j \\
    &= \sum_{S \in 2^n, i, j \in S} \alpha_S \left( \frac{1}{n} {n-1 \choose |S|-1}^{-1} \right)^2 
    + \sum_{S \in 2^n, i \notin S, j \in S} \alpha_S \left( \frac{1}{n} {n-1 \choose |S|-1}^{-1} \right) \left( -\frac{1}{n} {n-1 \choose |S|}^{-1} \right) \\
    & + \sum_{S \in 2^n, i \in S, j \notin S} \alpha_S \left( \frac{1}{n} {n-1 \choose |S|-1}^{-1} \right) \left( -\frac{1}{n} {n-1 \choose |S|}^{-1} \right) 
    + \sum_{S \in 2^n, i, j \notin S} \alpha_S \left( -\frac{1}{n} {n-1 \choose |S|}^{-1} \right)^2
\end{align}

Now if $\trainbudget \ge 2 \sum_{j=0}^p {n \choose j}$. We make $\alpha_S^{(j)} = 0$ for all $0 \le j \le p$ and $n-p \le j \le n$ (i.e., explicitly computing $\uf(S)$ for very small and very large $S$). 

Under this strategy, the property of $\Tilde{S}_n \Tilde{S}_n^T$ is very good in the sense that every diagonal entry has the same value, and every non-diagonal entry has the same value:
\begin{align}
    d_1 &= (\Tilde{S}_n)_i^T (\Tilde{S}_n)_i \\
    &= \sum_{S\in 2^n, i\in S, p+1 \le |S| \le n-p-1} \alpha_S \left( \frac{1}{n} {n-1 \choose |S|-1}^{-1} \right)^2 
    + \sum_{S\in 2^n, i\notin S, p+1 \le |S| \le n-p-1} \alpha_S \left( -\frac{1}{n} {n-1 \choose |S|}^{-1} \right)^2 \\
    &= \frac{1}{n^2} \sum_{k=p+1}^{n-p-1}
    \left( {n-1 \choose k-1}^{-1} + {n-1 \choose k}^{-1} \right)
\end{align}
If $p \ge 1$: 
\begin{align}
    d_2
    &= \frac{1}{n^2} \sum_{k=p+1}^{n-p-1} \left[
    {n-2 \choose k-2} {n-1 \choose k-1}^{-2} 
    -2 {n-2 \choose k-1} {n-1 \choose k-1}^{-1} {n-1 \choose k}^{-1} + {n-2 \choose k} {n-1 \choose k}^{-2}
    \right]
\end{align}
If $p = 0$: 
\begin{align}
    d_2 
    &= \frac{1}{n^2} \sum_{k=p+1}^{n-p-1} \left[
    {n-2 \choose k-2} {n-1 \choose k-1}^{-2} 
    -2 {n-2 \choose k-1} {n-1 \choose k-1}^{-1} {n-1 \choose k}^{-1} + {n-2 \choose k} {n-1 \choose k}^{-2}
    \right]\\
    &~~~ - \frac{2}{n(n-1)^2} \\
    &= d_2' - \frac{2}{n(n-1)^2}
\end{align}

Therefore we can write 
\begin{align}
    \Tilde{S}_n \Tilde{S}_n^T = (d_1-d_2) I_n + d_2 \mathbf{1}_n
\end{align}
where $\mathbf{1}_n$ is the matrix with 1 in all entries, which has rank 1 and thus the only non-zero eigenvalue is $n$. Therefore, the eigenvalues of $\Tilde{S}_n \Tilde{S}_n^T$ are $(d_1 - d_2 + nd_2)$ and $(d_1-d_2)$.

\textbf{Case of $p \ge 1$}:
\begin{align}
    &d_1 - d_2 + nd_2 \\
    &= d_1 + (n-1)d_2 \\
    &= \frac{1}{n^2} \sum_{k=p+1}^{n-p-1}
    \left( {n-1 \choose k-1}^{-1} + {n-1 \choose k}^{-1} \right) \\
    & + \frac{n-1}{n^2} \sum_{k=p+1}^{n-p-1} \left[
    {n-2 \choose k-2} {n-1 \choose k-1}^{-2} 
    -2 {n-2 \choose k-1} {n-1 \choose k-1}^{-1} {n-1 \choose k}^{-1} + {n-2 \choose k} {n-1 \choose k}^{-2}
    \right] \\
    &= \frac{1}{n^2} \sum_{k=p+1}^{n-p-1} \left[
    {n-1 \choose k-1}^{-1} + {n-1 \choose k}^{-1} 
    + (k-1) {n-1 \choose k-1}^{-1} - 2k {n-1 \choose k-1}^{-1} 
    + (n-k-1) {n-1 \choose k}^{-1}
    \right] \\
    &= \frac{1}{n^2} \sum_{k=p+1}^{n-p-1} \left[
    -k {n-1 \choose k-1}^{-1} + (n-k) {n-1 \choose k}^{-1}
    \right] \\
    &= 0
\end{align}
and 
\begin{align}
    &d_1-d_2 \\
    &= -nd_2 \\
    &= -\frac{1}{n} \sum_{k=p+1}^{n-p-1} \left[
    {n-2 \choose k-2} {n-1 \choose k-1}^{-2} 
    -2 {n-2 \choose k-1} {n-1 \choose k-1}^{-1} {n-1 \choose k}^{-1} + {n-2 \choose k} {n-1 \choose k}^{-2}
    \right] \\
    &= -\frac{1}{n(n-1)} \sum_{k=p+1}^{n-p-1} \left[
    (k-1) {n-1 \choose k-1}^{-1} 
    -2 k {n-1 \choose k-1}^{-1} + (n-1-k) {n-1 \choose k}^{-1}
    \right] \\
    &= -\frac{1}{n(n-1)} \sum_{k=p+1}^{n-p-1} \left[
    (-k-1) {n-1 \choose k-1}^{-1} + (n-1-k) {n-1 \choose k}^{-1}
    \right] \\
    &= -\frac{1}{n(n-1)} \sum_{k=p+1}^{n-p-1} \left[
    \frac{ -(k+1)(k-1)!(n-k)! + (n-1-k)k!(n-1-k)! }{(n-1)!}
    \right] \\
    &= -\frac{1}{n(n-1)} \sum_{k=p+1}^{n-p-1} \left[
    \frac{ -(k-1)!(n-k)! - k!(n-1-k)! }{(n-1)!}
    \right] \\
    &= \frac{1}{n(n-1)}\sum_{k=p+1}^{n-p-1} \left[ {n-1 \choose k-1}^{-1} + {n-1 \choose k}^{-1} \right] \\
    &\le \frac{2}{n(n-1)}\sum_{k=p}^{n-p-1}{n-1 \choose k}^{-1} \\
    &\le \frac{4}{n(n-1)}\sum_{k=p}^{n/2}{n-1 \choose k}^{-1} \\
    &\le \frac{4}{n(n-1)} {n-1 \choose p}^{-1} \sum_{k=p}^{n/2} \left(\frac{1}{2}\right)^{k-p} \label{eq:geometric} \\
    &\le \frac{8}{n(n-1)} {n-1 \choose p}^{-1} \\
    &\le \frac{8}{n(n-1)} \left(\frac{p}{n}\right)^p
\end{align}
where (\ref{eq:geometric}) is due to 
\begin{align}
{n-1 \choose k}^{-1} / {n-1 \choose k-1}^{-1}
= \frac{k! (n-1-k)!}{(k-1)!(n-k)!} = \frac{k}{n-k} \le 1/2
\end{align}
for $k \le n/2$. 
Therefore, when $1 \le p \le \log(n)$, we have 
\begin{align}
\norm{\Tilde{S}_n}_{op} = \sqrt{ \sigma_{\max} ( \Tilde{S}_n \Tilde{S}_n^T ) } \le \Tilde{O}\left(\frac{1}{n^{1 + p/2}}\right)
\end{align}

\textbf{Case of $p = 0$}:
For the first eigenvalue $d_1 + (n-1)d_2$, follow a similar argument as for the case of $p \ge 1$, we derive 
\begin{align}
    d_1 + (n-1)d_2 = 0
\end{align}
For the second eigenvalue $d_1 - d_2$, we have 
\begin{align}
    &d_1 - d_2\\
    &= -nd_2 \\
    &= -n \left( d_2' - \frac{2}{n(n-1)^2} \right) \\
    &= \frac{1}{n(n-1)}\sum_{k=2}^{n-2} \left[ {n-1 \choose k-1}^{-1} + {n-1 \choose k}^{-1} \right] + \frac{2}{(n-1)^2} \\
    &\le \frac{2}{(n-1)^2} + O \left(\frac{1}{n^3}\right)
\end{align}

Therefore, when $p=0$, i.e., only the full set and empty set's utility $\uf(\emptyset)$ and $\uf(N)$ is explicitly calculated, the worst-case L2 error guarantee can be improved to $\norm{\Tilde{S}_n}_{op} = \sqrt{ \sigma_{\max} ( \Tilde{S}_n \Tilde{S}_n^T ) } \le O\left(\frac{1}{n}\right)$. 
We also note that when $p = -1$, i.e., there are no clean estimation and $\hybridduf = \duf$, we have $\norm{\Tilde{S}_n}_{op} = \norm{S_n}_{op} \le O \left(\frac{1}{\sqrt{n}}\right)$ as derived in Theorem 3 of \cite{yan2020evaluating}. 
More generally, we have $\norm{\Tilde{S}_n}_{op} \le \Tilde{O}\left(\frac{1}{n^{1+p/2}}\right)$ for $-1 \le p \le \log(n)$ in general.

\section{Proof of Theorem \ref{thm:sv-perm}}

Recall that we say $\hat \phi$ is an $(\eps, \delta)$-approximation to the true SV in terms of $p$-norm if 
\begin{align}
    \Pr_{\hat \phi} \left[ \norm{\phi - \hat \phi}_p \le \eps \right] \ge 1-\delta
\end{align}
By Hoeffding, when $p=1$, we can achieve $(\eps, \delta)$-approximation with $m = O\left( \frac{n^2}{\eps^2} \log (\frac{n}{\delta}) \right)$. 

Similar to Appendix \ref{appendix:sv-infotheory}, since $\uf$ and $\duf$ can be viewed as a $2^n$-dimensional vector, we can construct a hybrid data utility vector $\hybridduf \in \R^{2^n}$ where the entry correspond to subset $S$ is $\uf(S)$ if it is within the clean samples, and $\duf(S)$ otherwise. 
For simplicity, we assume that both $m$ and $\evalbudget$ can be divided by $n$. 
Now consider the situation when the characteristic function $\hybridduf$ is noisy for $\evalbudget / n$ permutations among the $m/n$ total permutations. By triangle inequality, we have
\begin{align}
    \left| \hat \phi_i(\hybridduf) - \phi_i(v) \right|
    \le \left| \hat \phi_i(\hybridduf) - \hat \phi_i(v) \right| 
    + \left| \hat \phi_i(v) - \phi_i(v) \right|
\end{align}

The extra error term 
\begin{align}
    \left| \hat \phi_i(\hybridduf) - \hat \phi_i(v) \right| 
    &= \frac{1}{m/n} \sum_{j=1}^{\evalbudget/n} \left| \left[ 
    (\hat v(P_i^{\pi_j} \cup \{i\}) - \hat v(P_i^{\pi_j}) )
    - (v(P_i^{\pi_j} \cup \{i\}) - v(P_i^{\pi_j}))
    \right] \right| \\
    &\le \frac{1}{m/n} \sum_{j=1}^{\evalbudget/n} 
    \left[
    \left| \hat v(P_i^{\pi_j} \cup \{i\}) - v(P_i^{\pi_j} \cup \{i\}) \right| + \left| \hat v(P_i^{\pi_j}) - v(P_i^{\pi_j}) \right|
    \right]
\end{align}

Now fix the sampled permutations $\pi_1, \ldots, \pi_j$. Suppose the error distribution follows $|\hat v - v|/r \sim \D_r$ with the support of $2^n$ simplex, and smooth s.t. $\kappa_0 \le \Pr_{x \sim \D_r}[x] \le \kappa_1$. With this error distribution, the expected value of the extra error term can be bounded as 
\begin{align}
    \E_{\hat v - v \sim \D_r} 
    \left[ 
    \left| \hat \phi_i(\hybridduf) - \hat \phi_i(v) \right| 
    \right]
    &\le 
    \frac{1}{m/n} \sum_{j=1}^{\evalbudget/n}
    \left[
    \E_{\hat v - v \sim \D_r} \left[
    \left| \hat v(P_i^{\pi_j} \cup \{i\}) - v(P_i^{\pi_j} \cup \{i\}) \right| \right] \right. \\
    &~~~~~~~~~~~~+ \left. 
    \E_{\hat v - v \sim \D_r} \left[
    \left| \hat v(P_i^{\pi_j}) - v(P_i^{\pi_j}) \right| \right]
    \right] \\
    &\le \frac{\evalbudget}{m} \frac{2\kappa_1}{\kappa_0} \frac{r}{2^n} \label{eq:smoothguarantee} \\
    &= \gamma \frac{2\kappa_1}{\kappa_0} \frac{r}{2^n}
\end{align}
where the inequality in (\ref{eq:smoothguarantee}) is due to the following\footnote{This argument is inspired by \cite{yan2020evaluating}.}: let subset $C^{*}=\argmax_{C} \mathbb{E}_{\boldsymbol{v}-\hat{\boldsymbol{v}} \sim \mathcal{D}_{S_{r}}}\left[|v(C)-\hat{v}(C)|]\right.$ and subset $C^{\prime}=$ $\argmin_{C} \mathbb{E}_{\boldsymbol{v}-\hat{\boldsymbol{v}} \sim \mathcal{D}_{S_{r}}}[|v(C)-\hat{v}(C)|]$. 
\begin{align}
\mathbb{E}_{\boldsymbol{v}-\hat{\boldsymbol{v}} \sim \mathcal{D}_{S_{r}}}\left[\left|v\left(C^{*}\right)-\hat{v}\left(C^{*}\right)\right|\right]&=\int\left|v\left(C^{*}\right)-\hat{v}\left(C^{*}\right)\right| \operatorname{Pr}_{\mathcal{D}_{S_{r}}}(\boldsymbol{v}-\hat{\boldsymbol{v}}) d(\boldsymbol{v}-\hat{\boldsymbol{v}}) \\
& \leq \int\left|v\left(C^{*}\right)-\hat{v}\left(C^{*}\right)\right| \kappa_{1} d(\boldsymbol{v}-\hat{\boldsymbol{v}}) \\
& = \int\left|v\left(C^{\prime}\right)-\hat{v}\left(C^{\prime}\right)\right| \kappa_{1} d(\boldsymbol{v}-\hat{\boldsymbol{v}}) \label{eq:sym} \\
& \leq \int\left|v\left(C^{\prime}\right)-\hat{v}\left(C^{\prime}\right)\right| \frac{\kappa_{1}}{\kappa_{0}} \operatorname{Pr}_{\mathcal{D}_{S_{r}}}(\boldsymbol{v}-\hat{\boldsymbol{v}}) d(\boldsymbol{v}-\hat{\boldsymbol{v}}) \\
&=\frac{\kappa_{1}}{\kappa_{0}} \mathbb{E}_{\boldsymbol{v}-\hat{\boldsymbol{v}} \sim \mathcal{D}_{S_{r}}}\left[\left|v\left(C^{\prime}\right)-\hat{v}\left(C^{\prime}\right)\right|\right] \\
& \leq \frac{\kappa_{1}}{\kappa_{0}}\left(\frac{r}{2^{n}}\right) \label{eq:avg}
\end{align}

Here (\ref{eq:sym}) holds by symmetry as the expectation of any two vector coordinates under a uniform distribution over the simplex of vectors is the same. (\ref{eq:avg}) holds because every vector in the support of $\mathcal{D}_{S_{r}}$ has $\ell_1$ norm of $r$, $\sum_{C} \mathbb{E}_{\boldsymbol{v}-\hat{\boldsymbol{v}} \sim \mathcal{D}_{S_{r}}}[|v(C)-\hat{v}(C)|]=\mathbb{E}_{\boldsymbol{v}-\hat{\boldsymbol{v}} \sim \mathcal{D}_{S_{r}}}\left[\sum_{C}|v(C)-\hat{v}(C)|\right]=r$ and so by our choice of $C^{\prime}, \mathbb{E}_{\boldsymbol{v}-\hat{\boldsymbol{v}} \sim \mathcal{D}_{S_{r}}}\left[\left|v\left(C^{\prime}\right)-\hat{v}\left(C^{\prime}\right)\right|\right] \leq \frac{r}{2^{n}}$.

Integrating over $r$, we have 
\begin{align}
    \E_{\hat v - v \sim \D} 
    \left[ 
    \left| \hat \phi_i(\hat v) - \hat \phi_i(v) \right| 
    \right]
    \le \frac{\gamma}{2^{n-1}} \E_r\left[ \frac{\kappa_1(r)}{\kappa_0(r)}r \right]
\end{align}
Therefore
\begin{align}
    \E_{\hat v - v \sim \D} 
    \left[ 
    \left| \hat \phi_i(\hat v) - \phi_i(v) \right| 
    \right]
    \le \frac{\gamma}{2^{n-1}} \E_r\left[ \frac{\kappa_1(r)}{\kappa_0(r)}r \right] + 
    \left| \hat \phi_i(v) - \phi_i(v) \right| 
\end{align}

Denote $c = \frac{1}{2^{n-1}} \E_r\left[ \frac{\kappa_1(r)}{\kappa_0(r)}r \right]$. 

Hence 
\begin{align}
    &\Pr_{\pi_1, \ldots, \pi_m \sim Unif(\Pi)}\left[
    \E_{\hat v - v \sim \D} 
    \left[ 
    \norm{ \hat \phi(\hat v) - \phi(v) }_1
    \right] \le \eps + \gamma cn
    \right] \\
    &= 
    \Pr_{\pi_1, \ldots, \pi_m \sim Unif(\Pi)}\left[
    \gamma cn + 
    \sum_{i=1}^n
    \left| \hat \phi_i(v) - \phi_i(v) \right| 
     \le \eps + \gamma cn
    \right] \\
    &\ge 
    \Pr_{\pi_1, \ldots, \pi_m \sim Unif(\Pi)}\left[
    \sqcup_{i=1}^n
    \left| \hat \phi_i(v) - \phi_i(v) \right| 
     \le \eps/n
    \right] \\
    &\ge 1-\delta
\end{align}
whenever 
$$
m = O\left( \frac{n^2}{\eps^2} \log (\frac{n}{\delta}) \right)
$$
Even if $\gamma = \frac{\evalbudget}{m} \rightarrow 1$, as long as $c = o(\frac{1}{n})$, the irreducible error term $\gamma cn \rightarrow 0$ as $n \rightarrow \infty$. This is very possible since $c$ has a $\frac{1}{2^n}$ as a factor.

\section{Proof of Theorem \ref{thm:least-core}}
This proof directly extends the proof of Theorem 2 of \cite{yan2020ifyoulike}, which is based on the observation in \cite{balcan2015learning} and \cite{balkanski2017statistical} that estimating least core from finite samples is equivalent to the problem of learning an unknown linear function $(x, e)$ s.t. $\sum_{i \in S} x_i \ge v(S)$ for all $S \subseteq N$. Like \cite{yan2020evaluating} and \cite{balkanski2017statistical}, we use the following learnability result for linear classifiers \cite{shalev2014understanding}:

\begin{lemma}[Rademacher Complexity for Halfspaces]
Let $\mathcal{H}=\left\{\mathbf{w}:\|\mathbf{w}\|_{1} \leq B\right\}$ be the hypothesis class, and $\mathcal{Z}=\mathcal{X} \times \mathcal{Y}$ be the examples domain. Suppose $\mathcal{D}_{Z}$ is a distribution over $\mathcal{Z}$ s.t $\|\mathbf{x}\|_{\infty} \leq R$. Let the loss function $\ell: \mathcal{H} \times \mathcal{Z} \rightarrow \mathbb{R}$ be of the form $\ell(\mathbf{w},(\mathbf{x}, y))=\phi(\langle\mathbf{w}, \mathbf{x}\rangle, y)$ and $\phi: \mathbb{R} \times$ $\mathcal{Y} \rightarrow \mathbb{R}$ is such that for all $y \in \mathcal{Y}$, the scalar function $a \rightarrow \phi(a, y)$ is $\rho$-Lipschitz and such that $\max _{a \in[-B R, B R]}|\phi(a, y)| \leq c .$ Then for any $\Delta \in(0,1)$, with probability of at least $1-\Delta$ over the choice of an i.i.d. sample of size $m,\left(\mathbf{x}_{1}, y_{1}\right), \ldots,\left(\mathbf{x}_{m}, y_{m}\right)$ :
$$
\mathbb{E}_{(\mathbf{x}, y) \sim \mathcal{D}_{z}}[\ell(\mathbf{w},(\mathbf{x}, y))] \leq \frac{1}{m} \sum_{i=1}^{m} \ell\left(\mathbf{w},\left(\mathbf{x}^{i}, y^{i}\right)\right)+2 \rho B R \sqrt{\frac{2 \log (2 d)}{m}}+c \sqrt{\frac{2 \log (2 / \Delta)}{m}}
$$
for all $\mathbf{w} \in \mathcal{H}$.
\label{lemma:rademacher}
\end{lemma}

We assume that for every $S \subseteq N$, the error distribution has concentration property 
\begin{align}
    (1-\eps \sigma) v(S) \le \hat v(S) \le (1+\eps \sigma) v(S) 
    \label{eq:errordist-LC}
\end{align}
with probability at least $1-\mu$. If we have $\evalbudget$ noisy samples of $\hat v(S)$, then with probability at least $1- \evalbudget \mu$ that all of them satisfy (\ref{eq:errordist-LC}). 

We also require the following observation:
\begin{lemma}
For any $\eps, \sigma > 0$, $\delta < 1$, and $(x, e)$ computed from the linear program in (\ref{eq:leastcore-mc}) with each $\uf(S)$ replaced by $\hybridduf(S)$, 
\begin{align}
    \E_{S \sim \D} \left[ \left[ 1 - \frac{\sum_{i \in S}z_i + e}{v(S)} \right]_+ \right] \le \eps(\sigma + \delta) 
    \Rightarrow 
    \Pr_{S \sim \D} \left[ \sum_{i \in S} z_{i}+\estar+\eps(1+\sigma) \geq v(S) \right] \ge 1 - \sigma - \delta
\end{align}
\label{lemma:equivalence}
\end{lemma}

\begin{proof}
\begin{align}
    \left[1-\frac{\sum_{i \in S} z_{i}+e}{v(S)}\right]_{+} \le \eps 
    &\equi
    1-\frac{\sum_{i \in S} z_{i}+e}{v(S)} \le \eps \\
    &\equi \sum_{i \in S} z_{i}+e + \eps v(S) \ge v(S) \\
    &\Rightarrow \sum_{i \in S} z_{i}+e + \eps \ge v(S) \\
    &\Rightarrow \sum_{i \in S} z_{i}+ (e-\sigma \eps) + \eps + \sigma \eps \ge v(S) \\
    &\Rightarrow \sum_{i \in S} z_{i}+ \estar + \eps + \sigma \eps \ge v(S)
\end{align}

By Markov inequality, 
\begin{align}
    \Pr_{S \sim \D} \left[ \sum_{i \in S} z_{i}+\estar+\eps(1+\sigma) \geq v(S) \right] 
    &= \Pr_{S \sim \D} \left[ \left[1-\frac{\sum_{i \in S} z_{i}+e}{v(S)}\right]_{+} \le \eps \right] \\
    &\ge 1 - \frac{ \E_{S \sim \D} \left[ \left[1-\frac{\sum_{i \in S} z_{i}+e}{v(S)}\right]_{+} \right] }{\eps} \\
    &\ge 1 - \frac{\eps(\sigma + \delta)}{\eps} \\
    &\ge 1 - \sigma - \delta
\end{align}
\end{proof}

Turning back to the theorem's proof. First, we bound the $\ell_1$ norm of the solution solved from the linear program in (\ref{eq:leastcore-mc}). Suppose $(x, e)$ is the solution, then 
\begin{align}
\norm{(x, e)}_1 = v(N) + e \le v(N) + \max_S \hat v(S) \le v(N) + \max_S v(S) + \eps \sigma \le 2\max_S v(S) + \eps \sigma
\end{align}
where we stress that $v(N)$ is always explicitly computed in the Monte Carlo approach for Least core estimation. 

We take our hypothesis class to be 
\begin{align}
    \mathcal{H} = \{ z \in \R^{n+1}: \norm{z}_1 \le 2\max_S v(S) + \eps \sigma \}
\end{align}

For any data subset $S$, we define the corresponding data feature as $\mathbf{x}^{S}=\left(\frac{\mathbb{I}[i \in S]}{v(S)}, \frac{1}{v(S)}\right)$ and the label to be $y^{S}=1$. We define to $\mathcal{D}$ to be the uniform distribution over all $\left(\mathbf{x}^{S}, y^{S}\right)$ pairs. Next, suppose we obtain $m$ samples $S_{1}, \ldots, S_{m}$ from $\mathcal{D}$, we may run the linear program in $(\ref{eq:leastcore-mc})$ on the $m$ samples where $\evalbudget$ of the samples are replaced to $\mathbf{x}^{S}=\left(\frac{\mathbb{I}[i \in S]}{\duf(S)}, \frac{1}{\duf(S)}\right)$, which gives us a payoff allocation $\hat{z}$ and a value $\hat{e}$. We take our classifier to be of the form $\mathbf{w}=(\hat{\mathbf{z}}, \hat{e})$ and we define its loss $\ell$ to be:
$$
\begin{aligned}
\ell\left(\mathbf{w},\left(\mathbf{x}^{S}, y^{S}\right)\right) &=\ell\left((\hat{\mathbf{z}}, \hat{e}),\left(\left(\frac{\mathbb{I}[i \in S]}{v(S)}, \frac{1}{v(S)}\right), y^{S}\right)\right) \\
&=\left[y^{S}-(\hat{\mathbf{z}}, \hat{e}) \cdot\left(\frac{\mathbb{I}[i \in S]}{v(S)}, \frac{1}{v(S)}\right)\right]_{+} \\
&=\left[1-\frac{\sum_{i \in S} \hat{z}_{i}+\hat{e}}{v(S)}\right]_{+}
\end{aligned}
$$

Now, we utilize Lemma \ref{lemma:rademacher} with the remaining variables being $R=\frac{1}{\min _{S \neq \emptyset} v(S)}$, $B= 2 \max_{S}v(S) + \eps \sigma$, $\phi(a, y)=[y-a]_{+}$, $\rho=1$ and $c=1+2BR$. 
By definition of $\mathbf{x}^{S},\left\|\mathbf{x}^{S}\right\|_{\infty} \leq \frac{1}{\min _{S \neq \emptyset} v(S)}$. 
By definition of the hypothesis class, $\|(\mathbf{z}, e)\|_{1} \leq 2 \max _{S} v(S) + \eps \sigma$ for all $(\mathbf{z}, e) \in \mathcal{H}$, $\phi(a, y)=[y-a]_{+}$ is 1-Lipschitz. 
Since our example domain $\mathcal{Z}$ is such that $\mathcal{Y}=\{1\} .$ We may obtain upper bound $c$: $c=\max _{a \in[-B R, B R]}|\phi(a, y)|=\max _{a \in[-B R, B R]}[1-a]_{+} \leq(1--B R)=1+BR$. 

Now we bound the training error of $(x, e)$ on clean samples:
\begin{align}
    \frac{1}{m} \sum_{j=1}^{\evalbudget} \left[1-\frac{\sum_{i \in S_j} \hat{\mathbf{z}}_{i}+\hat{e}}{v(S_j)}\right]_{+} 
    \le \frac{1}{m} \sum_{j=1}^{\evalbudget} \left[1-\frac{\hat v(S_j)}{v(S_j)}\right]_{+} 
    \le \frac{1}{m} \sum_{j=1}^{\evalbudget} \left[1 - ( 1 - \eps \sigma) \right]_{+} 
    \le \frac{\evalbudget}{m} \eps \sigma
\end{align}

Therefore by Lemma \ref{lemma:rademacher}
\begin{align}
\mathbb{E}_{(\mathbf{x}, y) \sim \mathcal{D}}[l(\mathbf{w},(\mathbf{x}, y))] &=\mathbb{E}_{S \sim D}\left[\left[1-\frac{\sum_{i \in S} \hat{\mathbf{z}}_{i}+\hat{e}}{v(S)}\right]_{+}\right] \\
& \leq \frac{\evalbudget}{m} \eps \sigma + 2 \cdot 1 \cdot BR \sqrt{\frac{2 \log (2(n+1))}{m}}+(1+BR) \sqrt{\frac{2 \log (2 / \Delta)}{m}} \\
& \leq \frac{\evalbudget}{m} \eps \sigma + 2 BR \sqrt{\frac{2 \log (2(n+1))}{m}}+(1+BR) \sqrt{\frac{2 \log (2 / \Delta)}{m}} \\
& \leq \gamma \eps \sigma + 2 BR \sqrt{\frac{2 \log (2(n+1))}{m}}+(1+BR) \sqrt{\frac{2 \log (2 / \Delta)}{m}}
\end{align}

Using Lemma \ref{lemma:equivalence}, we need $m$ samples to be such that 
\begin{align}
    \gamma \eps \sigma + 2BR \sqrt{\frac{2 \log (2(n+1))}{m}}+(1+BR) \sqrt{\frac{2 \log (2 / \Delta)}{m}} \le \eps(\gamma \sigma + \delta)
\end{align}
and we get that 
$$
O\left(\frac{B^2 R^2\left(\log n+\log \left(\frac{1}{\Delta}\right)\right)}{\eps^{2} \delta^{2}}\right)
$$
samples suffice.

\section{Additional Experiment Details and Results}
\label{appendix:eval}

We show the implementation details (\ref{appendix:implementation}) and additional results on larger datasets valuation (\ref{appendix:data-removal}) here. 
For larger datasets, since it is impractical to compute the exact data value, we compare the performance of data value estimates on \emph{data removal} task, following existing data valuation literature \citep{ghorbani2019data,jia2019efficient, jia2019scalability, wang2020principled, yan2020ifyoulike}. Besides, we also evaluate the performance on data \emph{group} valuation (\ref{appendix:group-removal}), where the Shapley or least core values are assigned to a group of data rather than a single point.

\subsection{Implementation Details}
\label{appendix:implementation}

For tiny data experiment in Section 6, we use a small MLP model with 2 hidden layers as the data utility learning model, where the number of neurons in the hidden layers are 20 and 10, respectively. 
For the experiment of data valuation on larger data point / data group, we use MLP models with 3 hidden layers as the utility learning model. Each fully-connected layer has LeakyReLU as the activation function and is regularized by Dropout \citep{srivastava2014dropout}. 
The input datasets are encoded as binary vector in the natural sense, where $1$ indicates the data point is present, and $0$ indicates the data point is missing. 
We use Adam optimizer with learning rate $10^{-3}$, mini-batch size 32 to train all of the utility models mentioned above for up to 800 epochs. 

For fair comparisons, we always fix the same training budget $\trainbudget$ for different baselines. 
For Group Testing, we leverage half of the training budget to estimate the Shapley value of the last data point and the other half of the training budget to estimate the differences in Shapley value between data points. 
We use CVXOPT\footnote{\url{https://cvxopt.org/}} library to solve the constrained minimization problem in the least core approximation. 
We set the degree of CGA as 2 in the experiment. 
We use SGD optimizer with learning rate $10^{-3}$, batch size 32 to train the CGA model. 

For every experiment we conduct, we repeat each heuristic computation for 10 times to obtain the error bars.

\subsection{Additional Results on Tiny Dataset}
\label{appendix:more-on-tiny}

\begin{figure}
    \centering
    \includegraphics[width=\textwidth]{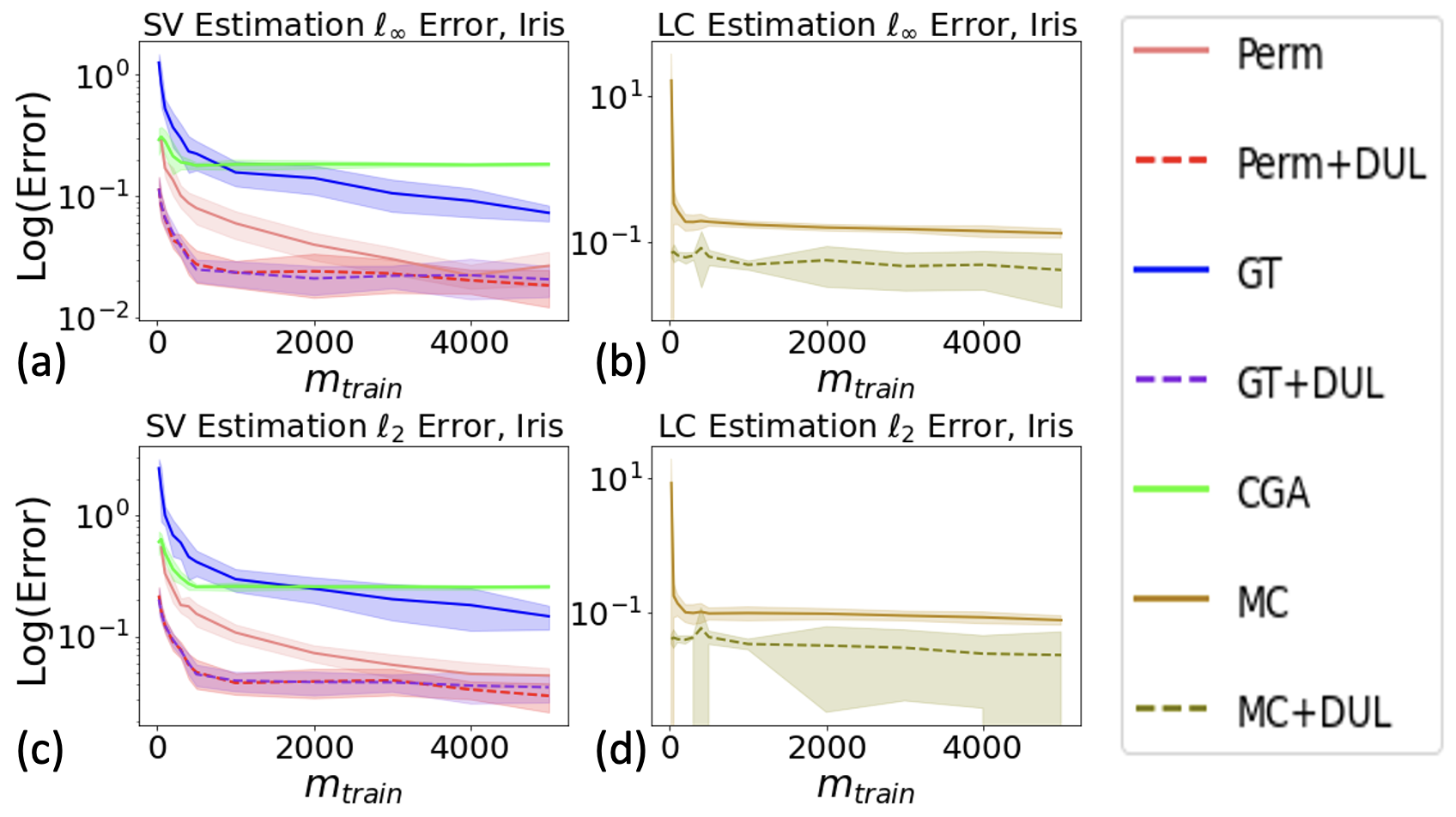}
    \caption{$\ell_2, \ell_\infty$ approximation error for Shapley and Least core with the change of number of samples for Iris dataset with 15 data points. DUL stands for data utility learning.
    }
    \label{fig:appendix-iris-error}
\end{figure}

We show $\ell_2, \ell_\infty$ errors in estimating SV or LC in Figure \ref{fig:appendix-iris-error}. For these two less stringent error metrics, we can still see that data utility learning still greatly reduces the Shapley/Least core estimation error with relatively small amount of sampled utilities. 

\begin{figure}
    \centering
    \includegraphics[width=\textwidth]{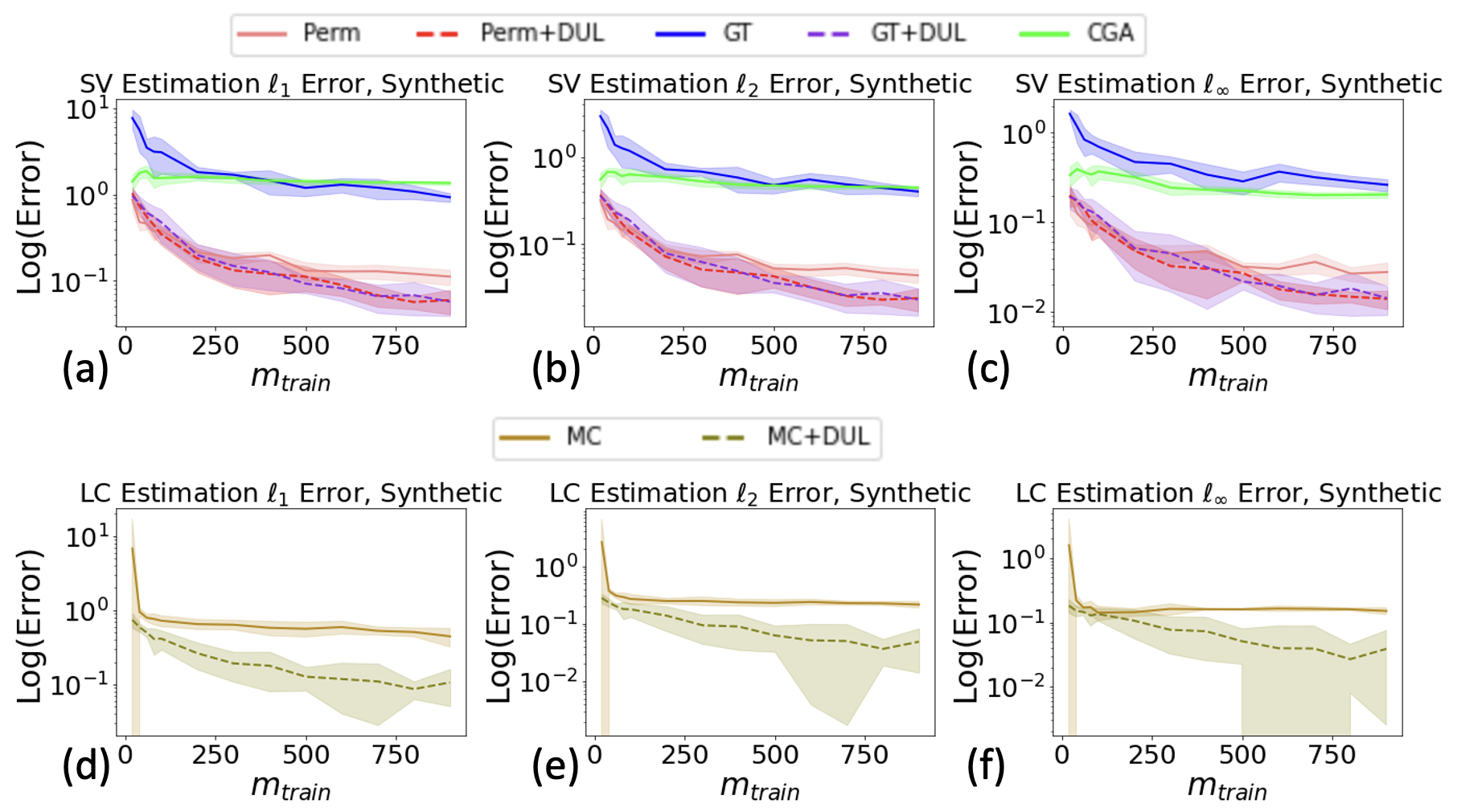}
    \caption{$\ell_1, \ell_2, \ell_\infty$ approximation error for Shapley and Least core with the change of number of samples for synthetic dataset with 10 data points. DUL stands for data utility learning.
    }
    \label{fig:appendix-synthetic-error}
\end{figure}

Moreover, we show additional results on a synthetic dataset in Figure \ref{fig:appendix-synthetic-error}. To generate the synthetic dataset, we sample 10 data points from a bivariate Gaussian distribution where the means are $0.1$ and $-0.1$ on each dimension, and covariance matrix is identity matrix. The labels are assigned based on the sign of the sum of the two features. We define the utility of a subset as the test accuracy of a logistic regression model trained on the subset. A logistic regression classifier trained on the 10 data points could achieve around 80\% test accuracy. 
As we can see from Figure \ref{fig:appendix-synthetic-error}, data utility learning again significantly improve the SV and LC estimation performance. 


\subsection{Data Removal }
\label{appendix:data-removal}

We evaluate the Shapley/Least core value estimations on larger datasets by comparing the performance on data removal task. Specifically, we remove the most (least) valuable portion of dataset and see how the utility of the remaining dataset changes. Intuitively, a better data value estimate can better identify the importance of data points. Hence, when the data with the highest (lowest) value estimates are removed, a better data value estimation method would lead to a faster (slower) performance drop. 
Similar to the experiment on tiny datasets, we experiment on both synthetic and real-world datasets. 
For the synthetic data generation, we sample 200 data points from a 50-dimensional Gaussian distribution, where the covariance matrix is identity matrix and the 50 mean parameters are sampled uniformly from $[-1, 1]$. Each data point is labeled by the sign of the sum of the data point vector. The utility of a subset is defined by the test accuracy of a logistic regression classifier trained on the dataset. 
For the real-world data experiment, we select 2000 data points from PubFig83 \citep{pinto2011scaling} dataset and the utility refers to the Top-5 accuracy of a simplified VGG model trained on it (for facial recognition). 
Since in error simulation experiment we showed that CGA does not perform well in estimating Shapley values for data valuation, and since CGA is not scalable to larger datasets as discussed in Section \ref{sec:dul-to-datavaluation}, we do not compare it as a baseline for this experiment. 

\begin{figure}[h]
    \setlength\belowcaptionskip{-10pt}
    \centering
    \includegraphics[width=\columnwidth]{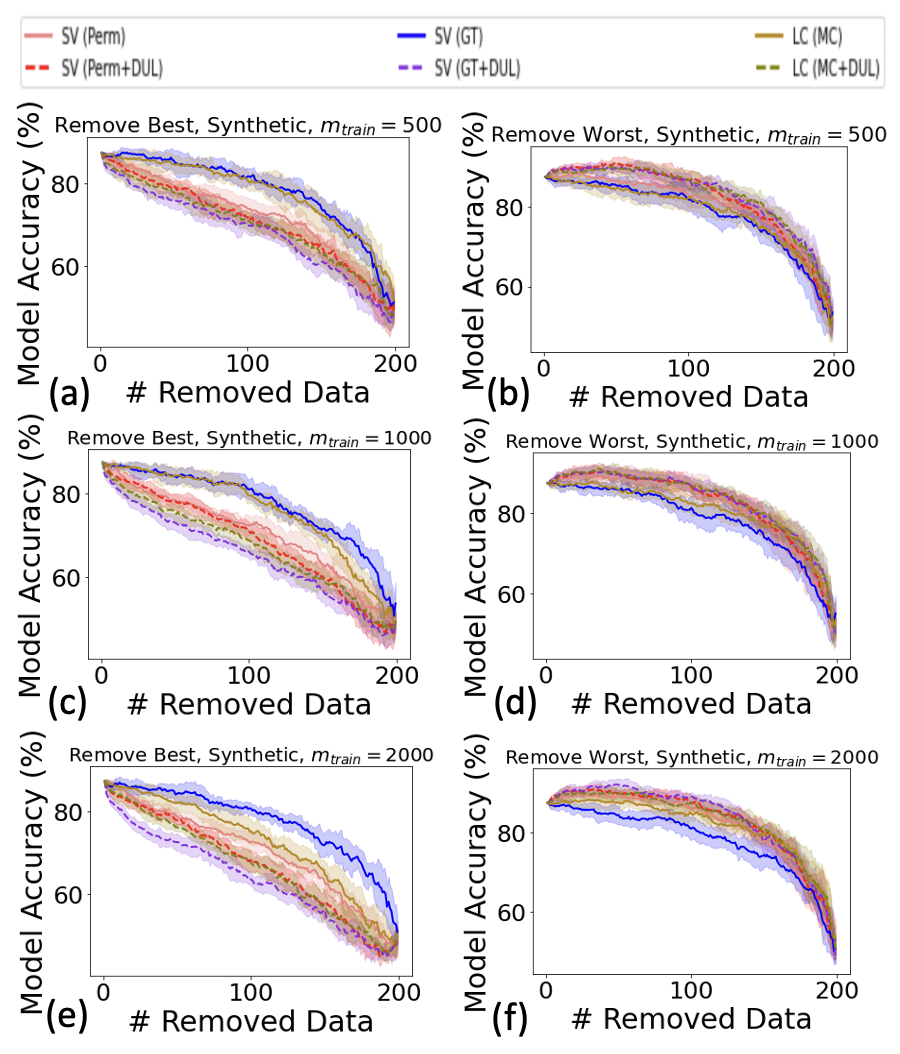}
    \caption{Curves of model test performance when the best/worst \textbf{synthetic} data points ranked according to Shapley (\textbf{SV}) or Least core (\textbf{LC}) estimations are removed. 
    The left column ((a), (c), (e)) removes the best data points. 
    The steeper the drop, the better. 
    The right column ((b), (d), (f)) removes the worst data points. 
    The sharper the rise, the better.
    DUL stands for data utility learning.
    }
    \label{fig:synthetic-full}
\end{figure}

\begin{figure}[h]
    \setlength\belowcaptionskip{-10pt}
    \centering
    \includegraphics[width=\columnwidth]{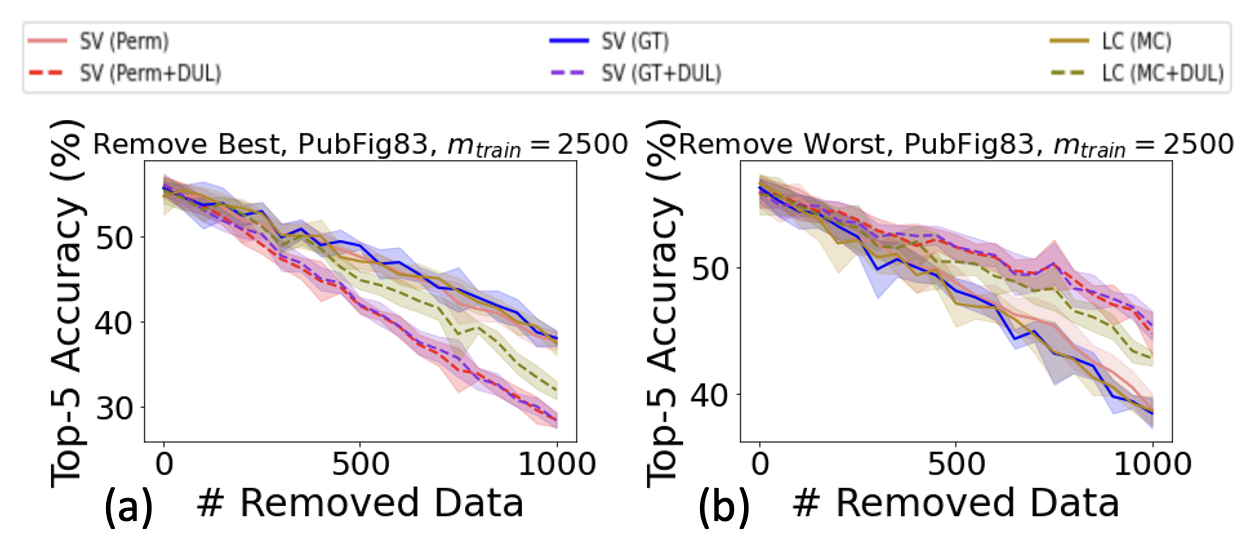}
    \caption{Curves of model test performance when the best/worst \textbf{PubFig83} data points ranked according to Shapley (\textbf{SV}) or Least core (\textbf{LC}) estimations are removed. 
    The left column ((a), (c), (e)) removes the best data points. 
    The steeper the drop, the better. 
    The right column ((b), (d), (f)) removes the worst data points. 
    The sharper the rise, the better.
    DUL stands for data utility learning.
    }
    \label{fig:pubfig-full}
\end{figure}

We experiment with different training budget $\trainbudget$, and we set the prediction budget $\evalbudget=10\trainbudget$ as utility model evaluation is much faster than retraining. 
We show the results of $\trainbudget=500, 1000, 2000$ for synthetic data in Figure \ref{fig:synthetic-full}, and $\trainbudget=2500$ for PubFig83 in Figure \ref{fig:pubfig-full}.  
These are clearly low-resource settings, as computing the exact Shapley or least core require $2^{200}$ times of training on synthetic data and $2^{2000}$ times of training on PubFig83. 
As we can see, the estimation heuristics equipped with data utility learning consistently performs better on identifying the most and least valuable data points.
This means that the Shapley and Least core values estimated with predicted utilities are at least more effective in predicting the most and least valuable (in a sense) data points in these settings. 
As a side note, the Shapley value estimated by Permutation sampling is superior to the Least core estimated by Monte-Carlo algorithm, which does not agree with the experiment results in \cite{yan2020ifyoulike}. 
An interesting future work is to better understand the comparison between Shapley and Least core for data valuation. 

\subsection{Data Valuation on Groups of Data}
\label{appendix:group-removal}

We also experiment on estimating Shapley and Least core values for groups of data points. This is a potentially more realistic and useful setting since in practice, more than one data records will be collected from one party. 
We divide Adult dataset \citep{adultdata} into 200 groups. The size of each group is varied. The proportion of data points allocated to each group is sampled from Dirichlet distribution where $\alpha$ is 30 in all dimensions. This design ensures that there are moderate amount of variations in group sizes. A utility sample in this setting refers to a coalition of groups and the performance of a logistic regression trained on the data points provided by the coalition. 

\begin{figure}[h]
    \setlength\belowcaptionskip{-10pt}
    \centering
    \includegraphics[width=\columnwidth]{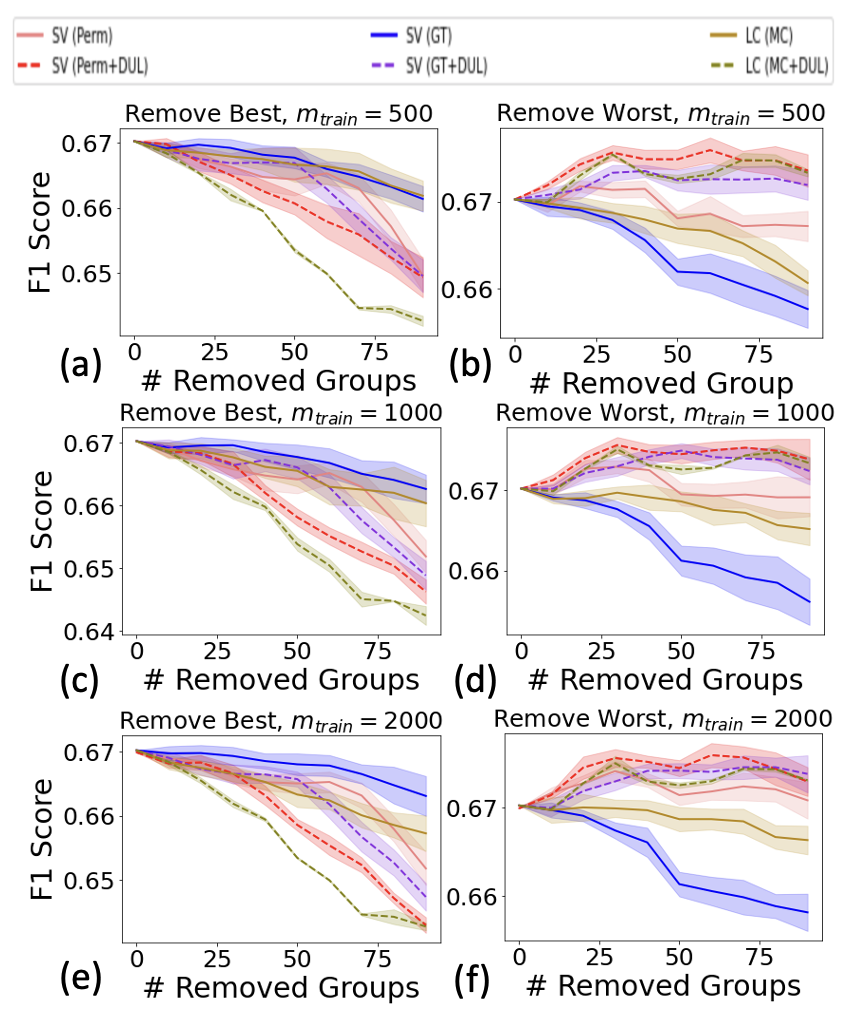}
    \caption{Curves of model test performance when the best/worst \textbf{adult} data groups ranked according to Shapley (\textbf{SV}) or Least core (\textbf{LC}) estimations are removed. 
    The left column ((a), (c), (e)) removes the best data points (groups). 
    The steeper the drop, the better. 
    The right column ((b), (d), (f)) removes the worst data points (groups). 
    The sharper the rise, the better.
    DUL stands for data utility learning.
    }
    \label{fig:adult-full}
\end{figure}

Since Adult dataset is a highly unbalanced dataset, we use F1-score as the utility metric. We vary the training budget $\trainbudget$ from 500 to 2000, and the prediction budget $\evalbudget = 10\trainbudget$ as in the previous section. 
As we can see from Figure \ref{fig:adult-full}, the heuristics equipped with data utility learning is again favorable to both of the Shapley and Least core estimation.


\end{document}